\documentclass[runningheads]{llncs}
\usepackage{graphicx}

\usepackage{tikz}
\usepackage{comment}
\usepackage{amsmath,amssymb} 
\usepackage{color}

\usepackage[accsupp]{axessibility}  

\usepackage{multirow}
\usepackage{booktabs}
\usepackage{subfigure}
\begin{document}
\pagestyle{headings}
\mainmatter

\title{Peng Cheng Object Detection Benchmark for Smart City} 


%
\author{Yaowei Wang\inst{1} \and
Zhouxin Yang\inst{1} \and
Rui Liu\inst{2} \and
Deng Li\inst{2} \and
Yuandu Lai \inst{2} \and
Leyuan Fang \inst{3} \and
Yahong Han \inst{2}
}
\authorrunning{Yaowei Wang et al.}
%
\institute{PengCheng Laboratory, Shenzhen, China \and College of Intelligence and Computing, Tianjin University, Tianjin, China \and  College of Electrical and Information Engineering, Hunan University, Changsha, China \\
\email{\{wangyw,yangzhx\}@pcl.ac.cn,\{ruiliu,lideng,yuandulai,yahong\}@tju.edu.cn}\\
\email{leyuan\_fang@hnu.edu.cn}}
\maketitle

\begin{abstract}
Object detection is an algorithm that recognizes and locates the objects in the image and has a wide range of applications in the visual understanding of complex urban scenes. 
Existing object detection benchmarks mainly focus on a single specific scenario and their annotation attributes are not rich enough, these make the object detection model is not generalized for the smart city scenes. Considering the diversity and complexity of scenes in intelligent city governance, we build a large-scale object detection benchmark for the smart city. Our benchmark contains about 500K images and includes three scenarios: intelligent transportation, intelligent security, and drones. For the complexity of the real scene in the smart city, the diversity of weather, occlusion, and other complex environment diversity attributes of the images in the three scenes are annotated. 
The characteristics of the benchmark are analyzed and extensive experiments of the current state-of-the-art target detection algorithm are conducted based on our benchmark to show their performance. Our benchmark is avaliable at \url{https://openi.org.cn/projects/Benchmark}.
\end{abstract}

\section{Introduction}

Many large-scale annotated visual datasets \cite{journals/ijcv/EveringhamGWWZ10,conf/eccv/LinMBHPRDZ14,conf/cvpr/GeigerLU12,conf/cvpr/CordtsORREBFRS16,conf/cvpr/YuCWXCLMD20} have driven the recent advances in multiple supervised learning tasks in computer vision significantly. A number of training examples are required to train deep learning models for achieving state-of-the-art performance on many tasks.

\begin{figure}[!h]
    \centering
    \includegraphics[width=\textwidth]{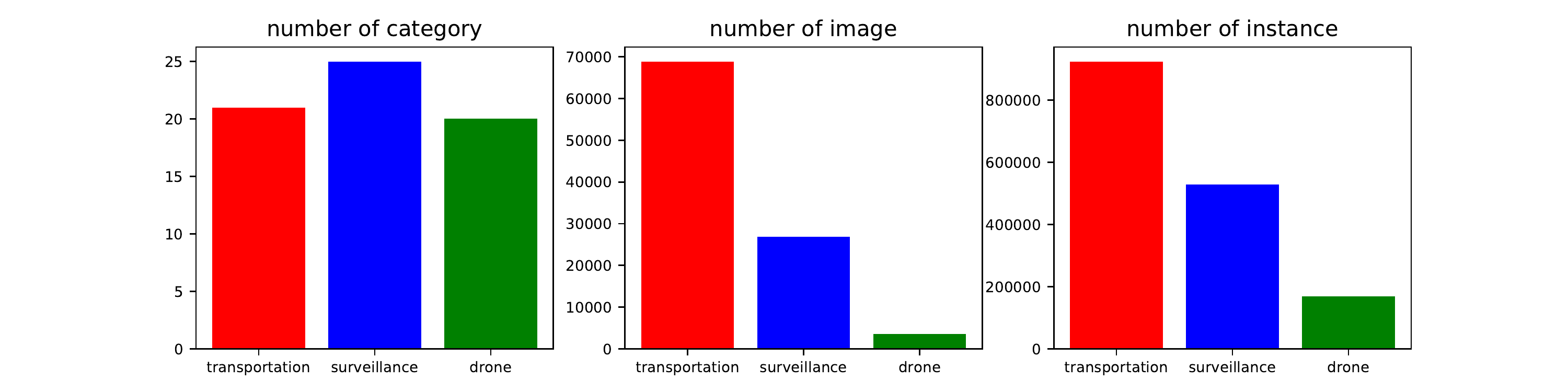}
    \caption{Statistics for the dataset.}
\end{figure}

\begin{figure}[!h]
    \centering
    \includegraphics[width=\textwidth]{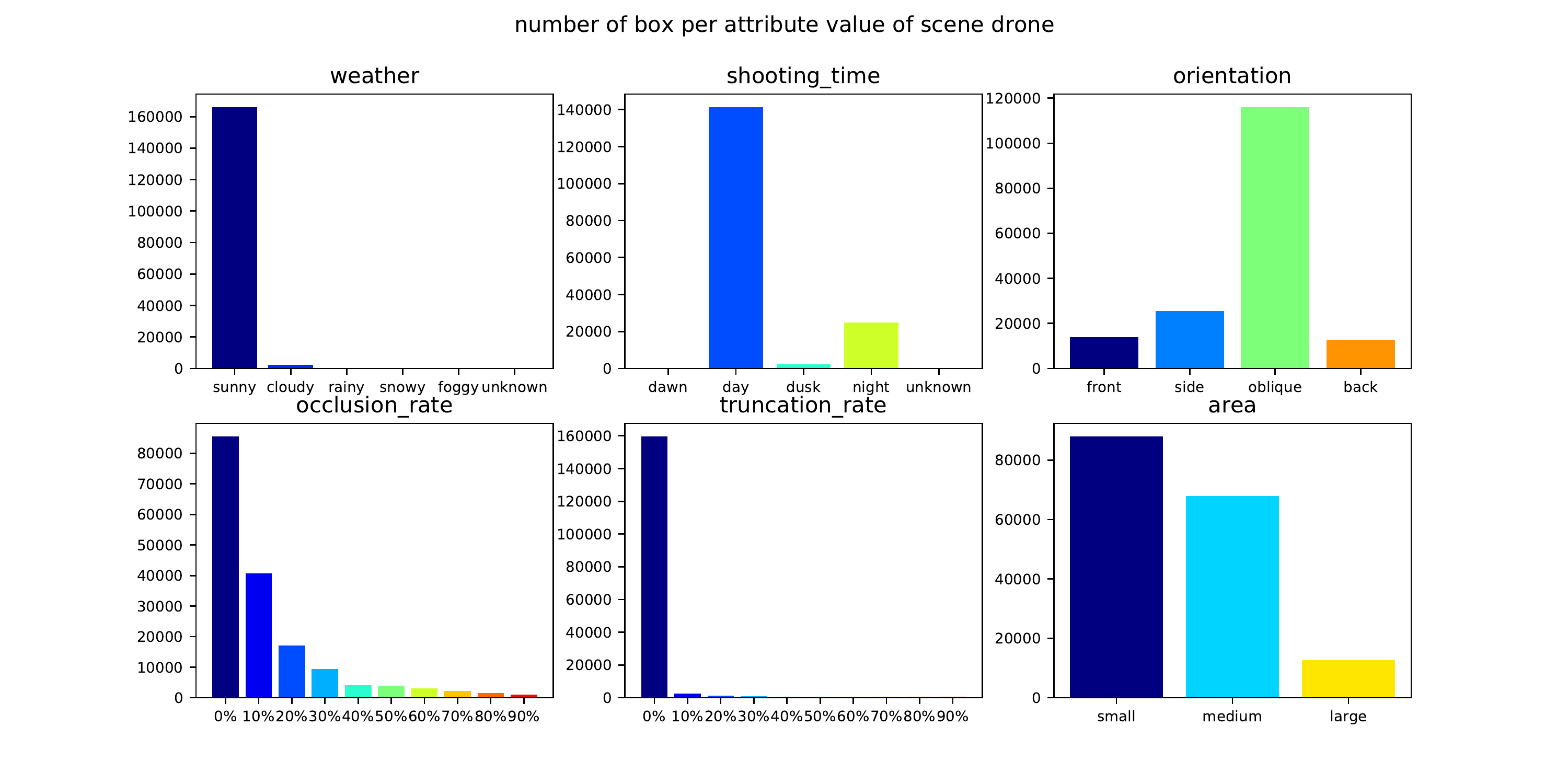}
    \caption{Number of box per attribute value of scene drone.}
\end{figure}

\begin{figure}[!h]
    \centering
    \includegraphics[width=\textwidth]{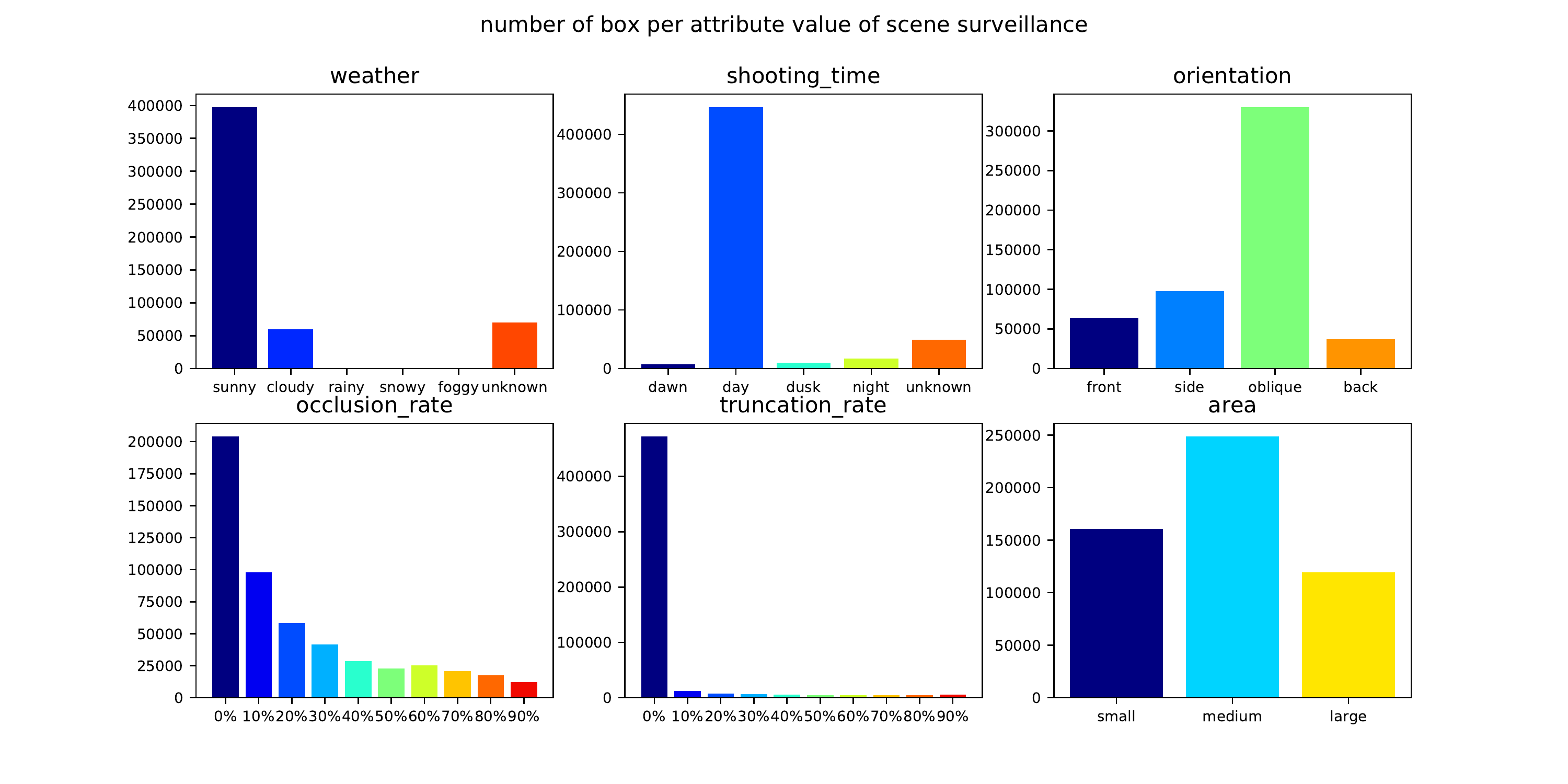}
    \caption{Number of box per attribute value of scene surveillance.}
\end{figure}

For the visual understanding of complex urban scenes, visual object detection algorithm, as an important branch of them, is used for a wide range of applications. And it is the basic support for high-level applications such as object tracking, behavior recognition, and video content analysis. Despite significant advances, it remains challenging in one or more important aspects when it comes to urban scenarios. Since they are complex, diverse, and flexible, such as object shape and size changes, orientation changes, illumination changes, weather changes, occlusion, blur, aggregation, background interference. From technological innovation to application in urban scenarios, visual object detection algorithms need to undergo rigorous testing benchmarks. Existing object detection algorithm test benchmarks mainly focus on a single specific scenario, only focus on the prediction performance of the algorithm, and lack a variety of test environments. Therefore, it is difficult for these test benchmarks to provide comprehensive and detailed test results, which are for the application of detection algorithms in urban scenes as a reference. In addition, the existing test benchmarks do not record and save the test environment and test process, which is not conducive to the traceability and promotion of tests.

In view of the deficiencies of existing test benchmarks, we aim to build a large-scale benchmark for complex urban scenes to overcome the limitations. We have been able to collect and annotate the dataset from the different urban scenes, consisting of over 500K diverse images. Our test benchmark has the following characteristics: covering more comprehensive urban scenarios; covering the test items in the learning process and prediction process; using a variety of test environments; recording the test operation process in an all-around way. And our benchmark introduces a variety of attributes including occlusion ratio, truncating ratio, illumination, direction, time period, and weather.

We conduct extensive evaluations of existing algorithms
on our new benchmarks. And our major contributions are as follows: 1) we build a large-scale object detection benchmark with diverse attributes for complex urban scenes; 2) extensive experiments of existing algorithms are conducted on this benchmark to show their performance; 3) our benchmark can support further study for computer vision in urban scenes.

\begin{figure}[!h]
    \centering
    \includegraphics[width=\textwidth]{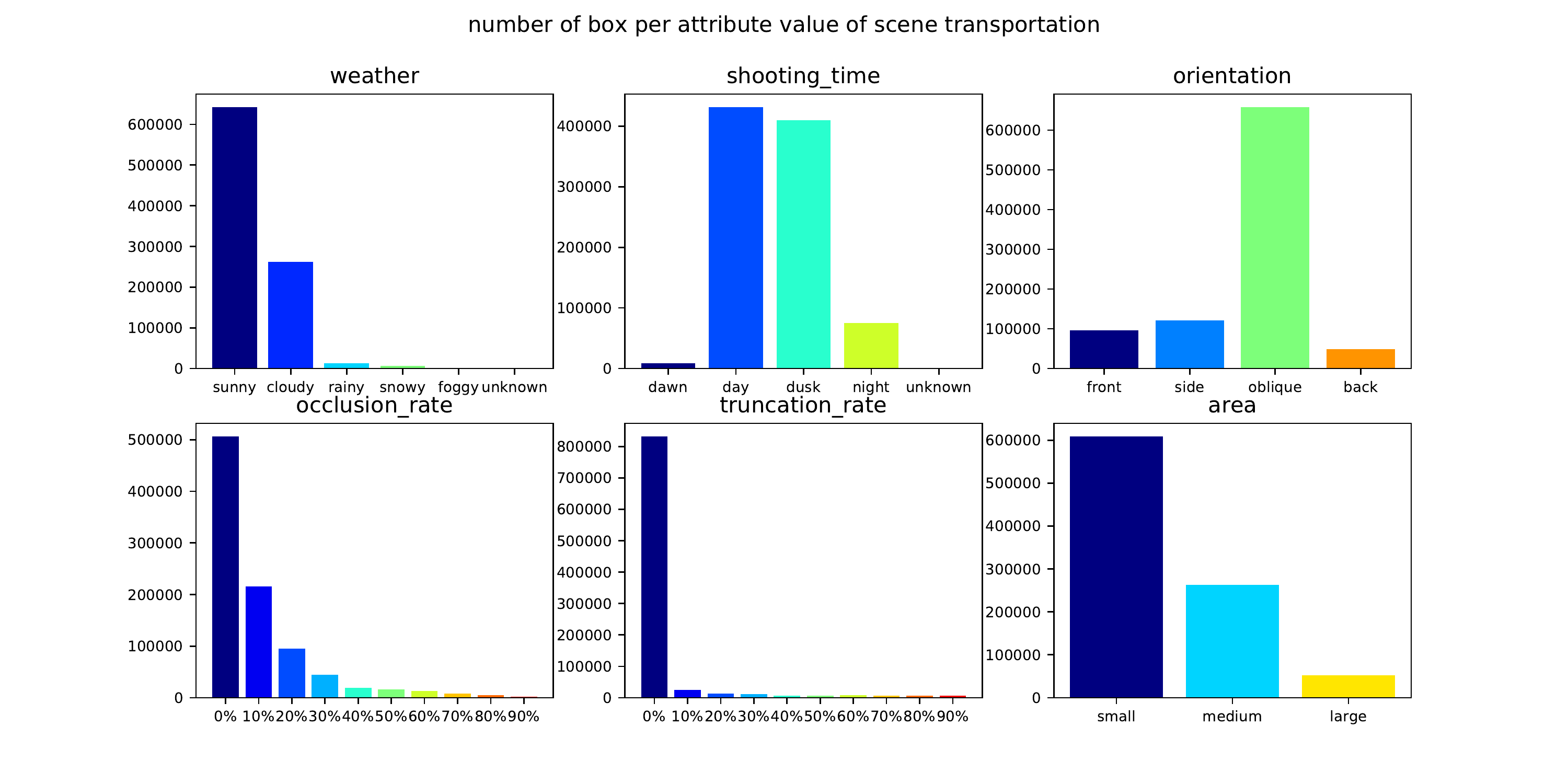}
    \caption{Number of box per attribute value of scene transportation.}
\end{figure}

\begin{figure}[!h]
    \centering
    \subfigure[]{\includegraphics[width=\textwidth]{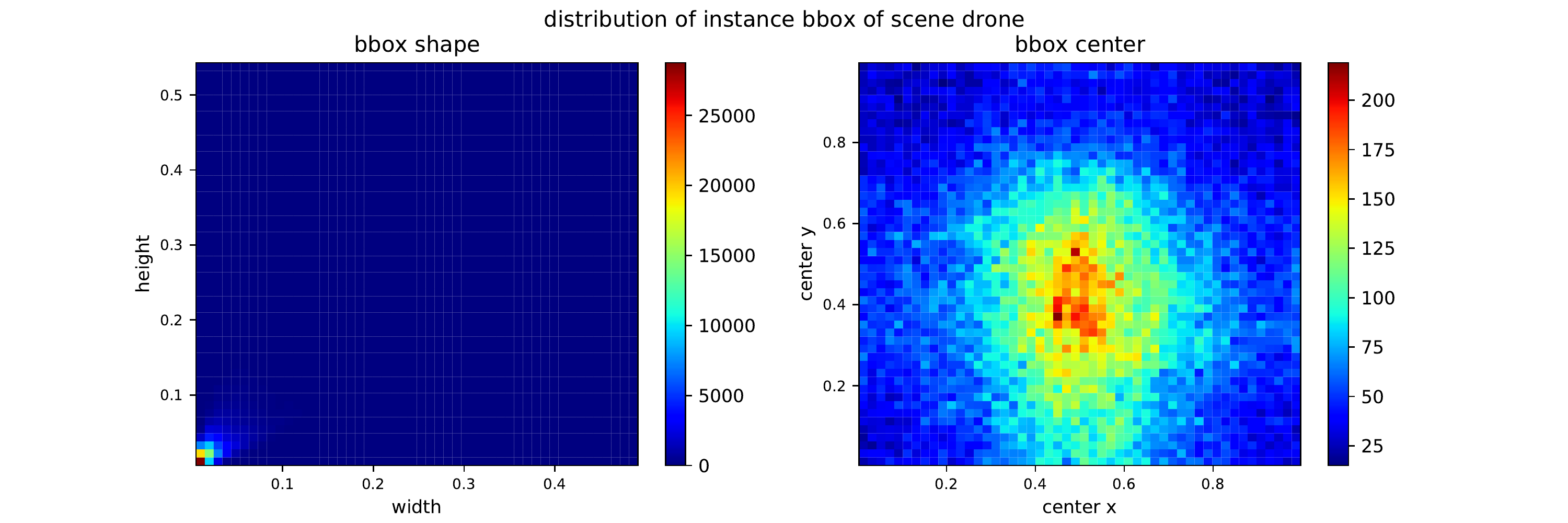}} \\
    \subfigure[]{\includegraphics[width=\textwidth]{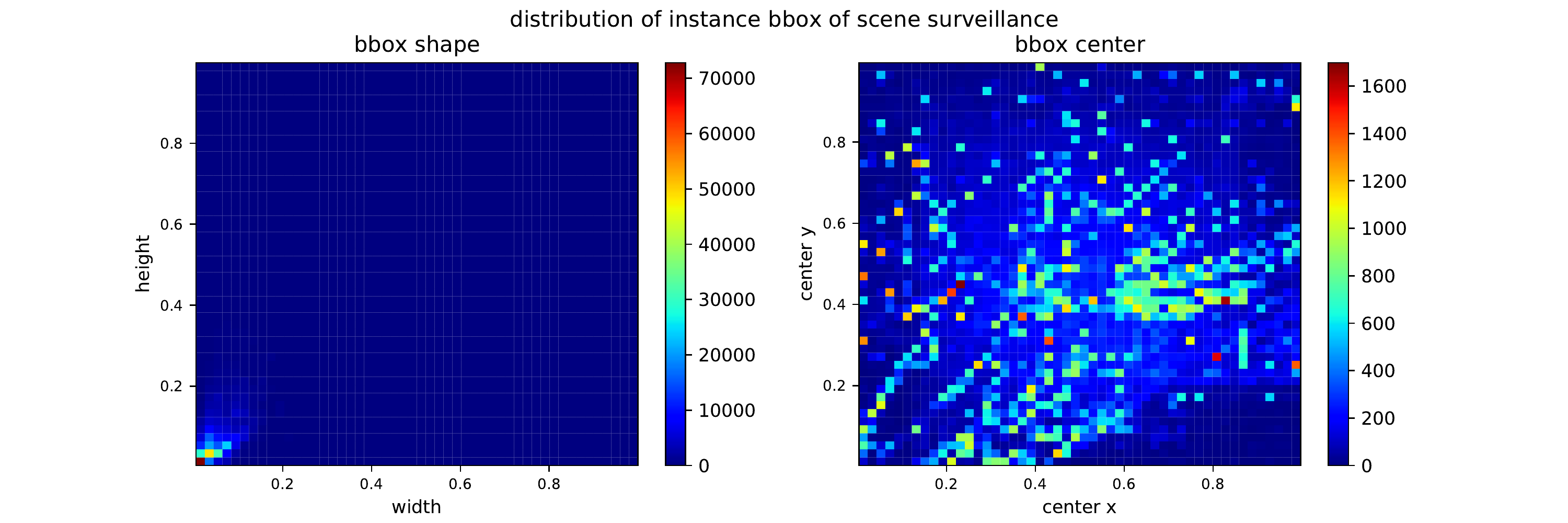}} \\
    \subfigure[]{\includegraphics[width=\textwidth]{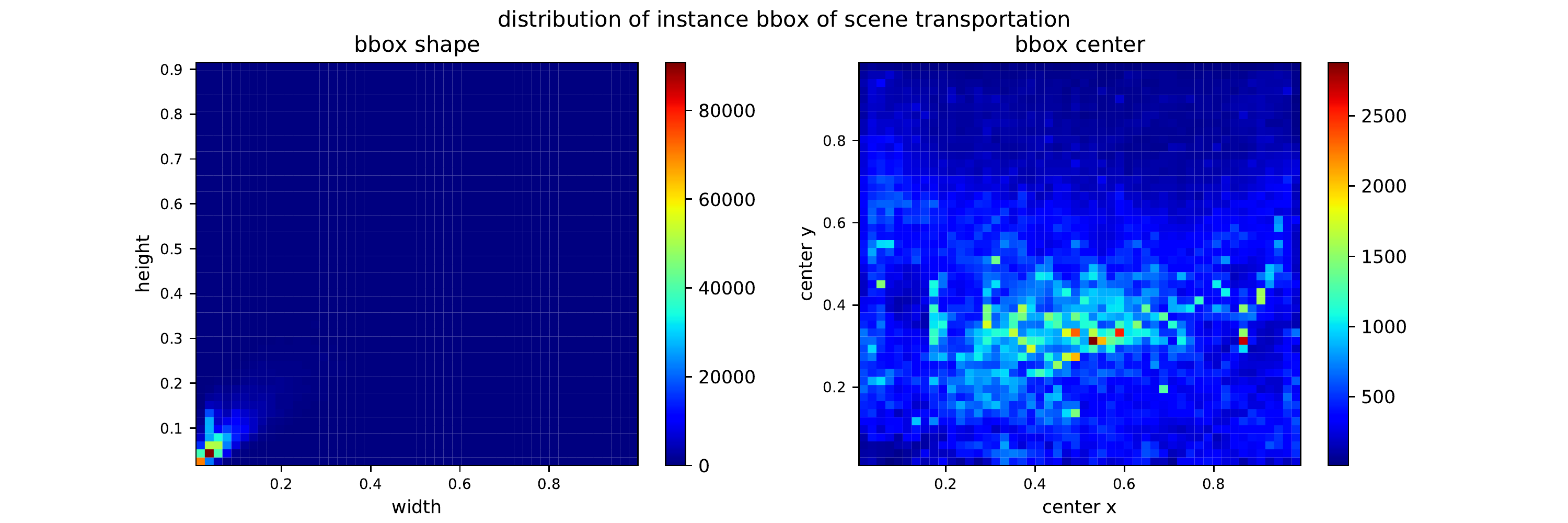}}
    
    \caption{Distribution of different bounding boxes.}
\end{figure}

\section{Dataset}

\subsection{Data Specifications}
In many scenarios of smart city governance, detection algorithms \cite{yang2016wider,lin2014microsoft,shao2019objects365,kuznetsova2020open,everingham2015pascal,neumann2018nightowls,zhu2020vision,yang2020advancing,nada2018pushing,zhang2017understanding} face great challenges brought by object shape and size changes, orientation changes, illumination changes, weather changes, occlusion, blur, aggregation, background interference, etc. Therefore, a more comprehensive dataset covering the scene is needed to truly reflect the comprehensive performance of the detection algorithm in the smart city governance scene.

Combined with the actual industrial application requirements of smart city governance, this benchmark constructs a dataset including intelligent transportation and intelligent security application scenarios. In terms of category selection, the dataset contains the categories that are common in these scenes and can reflect the challenges such as large object deformation, large aspect ratio, differences between large categories and differences within small categories. In addition, in order to facilitate users to deeply analyze the performance of the detection algorithm, this data set also provides multiple attribute annotations.

\begin{table}[t]
\scriptsize
\begin{center}
\caption{Comparison between our dataset and existing benchmark datasets. AS represents application scenario, IS represents intelligent security, IT represents intelligent transportation, NC represents number of categories, NS represents number of samples, NB represents total number of bounding boxes, NO represents number of objects, MI represents multiple illumination, MV represents multiple view, and MW represents multiple weather.} \label{tab:cap1}
\begin{tabular}{@{}l|ccccccccccc@{}}
\hline
\multicolumn{1}{c|}{} & AS     & NC & NS   & NB     & \begin{tabular}[c]{@{}c@{}}Size\\ (width)\end{tabular} & NO         & \begin{tabular}[c]{@{}c@{}}Object Width /\\ Sample Width\end{tabular} & MI & MV & MW & \begin{tabular}[c]{@{}c@{}}Attribute\\ Annotation\end{tabular}                                                      \\ \hline
VisDrone \cite{zhu2020vision}              & IS     & 11 & 9K   & 471K   & 2000                                                   & 1$\sim$100 & 1/100$\sim$1/10                                                       & $\checkmark$  &   &   & \begin{tabular}[c]{@{}c@{}}Occlusion, \\ Truncation\end{tabular}                                                    \\
MIO-TCD \cite{luo2018mio}               & IS     & 11 & 140K & 440K   & 200$\sim$480                                           & 1$\sim$10  & 1/10$\sim$1/2                                                         & $\checkmark$  &   & $\checkmark$  & N/A                                                                                                                 \\
Citycam \cite{zhang2017understanding}              & IT     & 10 & 60K  & 808K   & 352                                                    & 1$\sim$30  & 1/30$\sim$1/2                                                         & $\checkmark$  &   & $\checkmark$  & \begin{tabular}[c]{@{}c@{}}Orientation, \\ Weather\end{tabular}                                                     \\ \hline
Ours                  & IS, IT & 41 & 515K & 3,000K & 100$\sim$2000                                          & 1$\sim$100 & 1/100$\sim$1                                                          & $\checkmark$  & $\checkmark$  & $\checkmark$  & \begin{tabular}[c]{@{}c@{}}Occlusion, \\ Truncation, \\ Orientation, \\ Shooting \\ Period, \\ Weather\end{tabular} \\ \hline
\end{tabular}
\end{center}
\end{table}

Table \ref{tab:cap1} shows the comparison between this dataset and the existing benchmark datasets. Our dataset has a wider range of application scenarios, including intelligent security and intelligent transportation. In addition, our dataset has more categories, samples, objects, and bounding boxes. Moreover, compared with other datasets, our dataset has a wider range of image size and object width. More importantly, our dataset is the only one that contains multiple lights, multiple views and multiple weather conditions at the same time. Moreover, the annotations of other attributes are also more complete.

\begin{table}[t]
\small
\centering
\caption{The statistical data of our dataset. IS represents intelligent security, IT represents intelligent transportation, and NS represents number of samples.} \label{tab:cap2}
\begin{tabular}{@{}cccc@{}}
\hline
Scene               & NS                    & Source                                                                          & Quantity \\ \hline
\multirow{2}{*}{IT} & \multirow{2}{*}{250K} & Citycam \cite{zhang2017understanding}, UA-DETRAC \cite{wong2011patch}                                                              & 150K     \\ \cmidrule(l){3-4}
                    &                       & self collected                                                                  & 100K     \\ \cmidrule(l){3-4}
\multirow{2}{*}{IS} & \multirow{2}{*}{265K} & \begin{tabular}[c]{@{}c@{}}MIO-TCD \cite{luo2018mio}, VIRAT \cite{oh2011large}, \\ ChokePoint \cite{wong2011patch}, VISDrone \cite{zhu2020vision}\end{tabular} & 165K     \\ \cmidrule(l){3-4}
                    &                       & self collected                                                                  & 100K     \\ \hline
\end{tabular}
\end{table}

\begin{table}[t]
\tiny
\begin{center}
\caption{The categories and subclasses that our dataset contains.} \label{tab:cap3}
\begin{tabular}{@{}l|l@{}}
\hline
Categories      & \multicolumn{1}{c}{Subclasses}                                                                                                                                                                                                                                                                                                                                                                                                                                                                                                                                                                                                                                                                                                                                     \\ \hline
Traffic related & \begin{tabular}[c]{@{}l@{}}Passenger cars, cars, vans, caravans, off-road buses, special school buses, special buses, fence trucks,\\ multi-purpose trucks, van trucks, warehouse grid trucks, closed trucks, tank trucks, flat trucks,\\ container trucks, vehicle carriers, special structure trucks, dump trucks, special-purpose trucks,\\ semi-trailer tractors, full trailer tractors, Non cargo special operation vehicle, cargo special \\ operation vehicle, two wheeled motorcycle, three wheeled motorcycle, full trailer, semi-trailer,\\ wheel loading machinery, wheel excavation machinery, wheel leveling machinery, bicycle, electric \\ bicycle, motorcycle, tricycle, balance vehicle, skateboard, wheelchair, stroller, stroller.\end{tabular} \\ \hline
Human related   & Pedestrian, human face.                                                                                                                                                                                                                                                                                                                                                                                                                                                                                                                                                                                                                                                                                                                                            \\ \hline
Clothing        & Coat, pants, shoes.                                                                                                                                                                                                                                                                                                                                                                                                                                                                                                                                                                                                                                                                                                                                                \\ \hline
Appendage       & Hat, scarf, umbrella, helmet, glasses, mask, watch, mobile phone, backpack, handbag, trolley case.                                                                                                                                                                                                                                                                                                                                                                                                                                                                                                                                                                                                                                                                 \\ \hline
\end{tabular}
\end{center}
\end{table}

\begin{table}[t]
\tiny
\begin{center}
\caption{Object size division, occlusion ratio division, and truncation ratio division. Where, $W$, $R_o$, and $R_t$ indicate the width of the object, the ratio of the occlusion, and the ratio of the truncation, respectively.} \label{tab:cap4}
\begin{tabular}{@{}c|ccc|ccc|c@{}}
\cmidrule(r){1-2} \cmidrule(lr){4-5} \cmidrule(l){7-8}
\begin{tabular}[c]{@{}c@{}}Object Size\end{tabular}   & \multicolumn{1}{c}{Description}                                                                                  &  & \begin{tabular}[c]{@{}c@{}}Occlusion Ratio\end{tabular}  & \multicolumn{1}{c}{Description}                                                                    &  & \begin{tabular}[c]{@{}c@{}}Truncation Ratio\end{tabular}  & Description                                                                                         \\ \cmidrule(r){1-2} \cmidrule(lr){4-5} \cmidrule(l){7-8}
\begin{tabular}[c]{@{}c@{}}Small\end{tabular}  & \begin{tabular}[c]{@{}l@{}}$W$  $<$ 32 pixels \end{tabular}                                     &  & \begin{tabular}[c]{@{}c@{}}No \end{tabular}     & \begin{tabular}[c]{@{}l@{}}$R_o=$ 0\%\end{tabular}                              &  & \begin{tabular}[c]{@{}c@{}}No \end{tabular}     & \begin{tabular}[c]{@{}l@{}}$R_t =$0\% \end{tabular}                              \\ \cmidrule(r){1-2} \cmidrule(lr){4-5} \cmidrule(l){7-8}
\begin{tabular}[c]{@{}c@{}}Medium\end{tabular} & \begin{tabular}[c]{@{}l@{}}32 pixels$\leq$ $W$  $<$ 96 pixels\end{tabular} &  & \begin{tabular}[c]{@{}c@{}}Mild\end{tabular}   & \begin{tabular}[c]{@{}l@{}}$R_o <$   30\%\end{tabular}                   &  & \begin{tabular}[c]{@{}c@{}}Mild\end{tabular}   & \begin{tabular}[c]{@{}l@{}}$R_t <$   30\%\end{tabular}                   \\ \cmidrule(r){1-2} \cmidrule(lr){4-5} \cmidrule(l){7-8}
\begin{tabular}[c]{@{}c@{}}Large\end{tabular}  & \begin{tabular}[c]{@{}l@{}}96 pixels$\leq$ $W$\end{tabular}                              &  & \begin{tabular}[c]{@{}c@{}}Severe\end{tabular} & \begin{tabular}[c]{@{}l@{}}$R_o \geq$ 30\%\end{tabular} &  & \begin{tabular}[c]{@{}c@{}}Severe\end{tabular} & \begin{tabular}[c]{@{}l@{}}$R_t \geq$ 30\%\end{tabular} \\ \cmidrule(r){1-2} \cmidrule(lr){4-5} \cmidrule(l){7-8}
\end{tabular}
\end{center}
\end{table}


\begin{table}[!h]
\begin{center}
\caption{Other attributes include orientation, shooting period, and weather.} \label{tab:cap5}
\begin{tabular}{@{}c|l@{}}
\hline
\begin{tabular}[c]{@{}c@{}}Other\\ Attributes\end{tabular} & \multicolumn{1}{c}{Dscription}     \\ \hline
Orientation                                                & Front, side, oblique side, back   \\
Shooting period                                            & Day, night, early morning/evening \\
Weather                                                    & Sunny, overcast, rain, snow, fog  \\ \hline
\end{tabular}
\end{center}
\end{table}

\begin{table}[!h]
\scriptsize
\centering
\caption{Comparison between different models on intelligent transportation scene.}
\begin{tabular}{l|cccccc|cccccc}
\hline
Model        & mAP & AP@50 & AP@75 & AP@s & AP@m & AP@l & mAR & AR@50 & AR@75 & AR@s & AR@m & AR@l \\ \hline
YOLO-v4-tiny &19.8  &33.9  &19.6 & 15.6   &26.1  &26.7  &33.0  & 58.2 &31.9       & 29.2	   &26.7  & 45.2      \\
YOLO-v4      &31.6  &49.1  &34.1       &28.2    &36.7  & 37.2&51.3  &79.5  &54.8       & 30.5   & 40.6 & 57.5      \\
CenterNet    &25.0  & 45.0 & 23.6      &21.5	    & 28.4 & 28.4 &41.8  & 73.5 & 40.4      &38.6    & 46.8	 &47.0      \\
Faster R-CNN  & 32.6 &53.2  & 34.5      & 29.5   &35.6  & 40.4 &45.3  & 45.3 & 45.3      &42.6    &43.8  & 46.0     \\
DETR & 0.293 & 50.8 & 29.0      &25.8    &346  &42.6 & 24.3 &412  & 42.9      & 39.8   & 42.4 & 49.6      \\
Deformable &34.0  & 55.7 & 36.2  & 30.9   &37.1  &46.4 &28.3  &48.0  &49.9       & 47.7  & 48.1 &73.0      \\ \hline
\end{tabular}
\label{i_transportation}
\end{table}

\begin{table}[!h]
\scriptsize
\centering
\caption{Comparison between different models on intelligent surveillance scene.}
\begin{tabular}{l|cccccc|cccccc}
\hline
Model        & mAP & AP@50 & AP@75 & AP@s & AP@m & AP@l & mAR & AR@50 & AR@75 & AR@s & AR@m & AR@l \\ \hline
YOLO-v4-tiny & 16.6 &  31.0      &15.3    &3.9  &16.8 & 22.1 &27.3  &49.9       &25.3    & 11.9 &31.4 & 36.2      \\
YOLO-v4      & 25.3 &41.8  &25.8       &11.0    &27.7  &30.6 &40.6  &66.8  & 41.0      &25.8    & 44.1 &47.2       \\
CenterNet    &21.6  &39.3  & 21.0      & 13.5   & 24.3 &25.6 &37.6  & 64.9 & 37.3  &29.7    &40.1    &42.3   \\
Faster R-CNN   &29.9  &51.4  & 29.4      & 18.2   &33.0  &41.5  &39.4  & 39.4 &  39.4     &27.3    &42.8  & 50.4    \\
DETR &0.294  & 54.8 & 26.9      & 16.6   &30.8  & 50.1 & 24.1 &39.4  & 42.4     & 29.4   & 44.0 & 60.6    \\
Deformable &32.4  &54.2  & 32.8      & 20.5   & 34.7 & 45.6 &26.8  &44.1  &46.8       &  34.9  & 49.7 &59.5     \\ \hline
\end{tabular}
\label{i_surveillance}
\end{table}

\begin{table}[!h]
\scriptsize
\centering
\caption{Comparison between different models on drone scene.}
\begin{tabular}{l|cccccc|cccccc}
\hline
Model        & mAP & AP@50 & AP@75 & AP@s & AP@m & AP@l & mAR & AR@50 & AR@75 & AR@s & AR@m & AR@l \\ \hline
YOLO-v4-tiny & 4.6	 & 8.3 & 46 & 0.6 &5.5   &7.6  & 10.4 &199	  &9.7	  &3.0       &15.2	      & 16.8      \\
YOLO-v4      & 6.1 & 1.1 &6.0      &1.7   &9.6  & 11.2 &15.0 	  &26.3  & 14.7     & 8.5   &20.6  &21.6       \\
CenterNet    & 9.4 & 16.5 &9.1       &6.9    &10.2  & 10.0 &16.2	  &27.6  & 16.1      &13.8    & 16.7 & 16.5      \\
Faster R-CNN  & 10.6 &19.3  & 10.4      & 5.0   &15.9  & 24.5 & 17.4 & 17.4 &  17.4     &  09.6  & 25.6 &34.6     \\
DETR & 0.056 & 12.2 & 4.0      &1.6    &8.1  &19.0 &3.1  & 8.0 &  9.8     &  4.3  & 14.3 & 25.4     \\
Deformable &9.3  &17.4  &8.8       &4.4    &13.1  & 22.6 & 5.3 &14.1  & 16.9   &  10.0  & 22.9 & 35.7    \\ \hline
\end{tabular}
\label{i_drone}
\end{table}

\begin{table}[!h]
\footnotesize
\centering
\caption{Test Environment in the learning stage.}
\begin{tabular}{cccccc}
\hline
Category & Calculation Unit     & Memory & Framework                                                                        & Toolkit                                              & Hashrate   \\ \hline
GPU      & Tesla V100           & 256GB  & \begin{tabular}[c]{@{}c@{}}Pytorch 1.5\\ Tensorflow 2.2\\ Caffe 1.0\end{tabular} & \begin{tabular}[c]{@{}c@{}}CUDA 10.2\\ CUDNN 7.5\end{tabular} & 256 TFLOPS (FP16) \\ \hline
ASIC     & Ascend 910           & 32GB   & MindSpore 0.6.0                                                                  & -                                                    & 286 TFLOPS (FP16) \\ \hline
\end{tabular}
\end{table}

\begin{table}[!h]
\scriptsize
\centering
\caption{Test environment in the prediction stage.}
\begin{tabular}{ccccccc}
\hline
Category              & Calculation Unit & Memory & Framework                                                                                         & Toolkit                                                                        & Hashrate         & Power \\ \hline
\multirow{3}{*}{GPU}  & Tesla T4         & 16GB   & \multirow{3}{*}{\begin{tabular}[c]{@{}c@{}}Pytorch 1.5\\ Tensorflow 2.2\\ Caffe 1.0\end{tabular}} & \multirow{3}{*}{\begin{tabular}[c]{@{}c@{}}CUDA 10.2\\ CUDNN 7.5\end{tabular}} & 65 TFLOPS(FP16)  & 70W               \\
                      & NVIDIA V100    & 32GB    &                                                                                                   &                                                                                & 256 TFLOPS(FP16) & 10W               \\
                      & NVIDIA Xavier   & 4GB    &                                                                                                   &                                                                                & 21 TFLOPS(FP16) & 15W               \\ \hline
ASIC & Ascend 310       & 32GB   & MindSpore 0.6.0                                                                                   & -                                                                              & 8 TFLOPS(FP16)   & 5.5W   \\ \hline
\end{tabular}
\end{table}

\begin{table}[!h]
\footnotesize
\centering
\caption{Test items in the learning stage.}
\begin{tabular}{ccc}
\hline
  & Indicator & Explanation \\ \hline
1 & ${CT}_{env=x}$  & The learning time when the learning test environment is $x$. \\
2 & ${Epochs}_{env=x}$ & The total period of learning when the learning test environment is $x$.    \\ \hline
\end{tabular}
\end{table}

\begin{table}[!h]
\footnotesize
\centering
\caption{Test items (Precision) in the prediction stage.}
\begin{tabular}{ccc}
\hline
  & Indicator & Explanation \\ \hline
1 & $P^{IOU=0.5}_{cat=x}$         & Precision of $x$ class when IOU is 0.5 and class confidence threshold is 0.5             \\
2 & $P^{IOU=0.75}_{cat=x}$        & Precision of $x$ class when IOU is 0.75 and class confidence threshold is 0.5             \\
3 & $P^{IOU=0.5}$         &    Mean precision when IOU is 0.5 and class confidence threshold is 0.5       \\
4 & $P^{IOU=0.75}$         &    Mean precision when IOU is 0.75 and class confidence threshold is 0.5             \\ \hline
\end{tabular}
\end{table}

\begin{table}[!h]
\footnotesize
\centering
\caption{Test items (Recall) in the prediction stage.}
\begin{tabular}{ccc}
\hline
  & Indicator & Explanation \\ \hline
1 & $R^{IOU=0.5}_{cat=x}$         & Recall of $x$ class when IOU is 0.5 and class confidence threshold is 0.5            \\
2 & $R^{IOU=0.75}_{cat=x}$         & Recall of $x$ class when IOU is 0.75 and class confidence threshold is 0.5             \\
3 & $R^{IOU=0.5}$         & Mean precision when IOU is 0.5 and class confidence threshold is 0.5            \\
4 & $R^{IOU=0.75}$         & Mean precision when IOU is 0.75 and class confidence threshold is 0.5             \\ \hline
\end{tabular}
\end{table}

\begin{table}[!h]
\footnotesize
\centering
\caption{Test items (F1) in the prediction stage.}
\begin{tabular}{ccc}
\hline
  & Indicator & Explanation \\ \hline
1 & ${F1}^{IOU=0.5}_{cat=x}$        & F1 of $x$ class when IOU is 0.5 and class confidence threshold is 0.5             \\
2 & ${F1}^{IOU=0.75}_{cat=x}$        & F1 of $x$ class when IOU is 0.75 and class confidence threshold is 0.5            \\
3 & ${F1}^{IOU=0.5}$         & Mean F1 when IOU is 0.5 and class confidence threshold is 0.5            \\
4 & ${F1}^{IOU=0.75}$         & Mean F1 when IOU is 0.75 and class confidence threshold is 0.5          \\ \hline
\end{tabular}
\end{table}

\begin{table}[!h]
\small
\centering
\caption{Test items (Average Precision) in the prediction stage.}
\begin{tabular}{ccc}
\hline
  & Indicator & Explanation \\ \hline
1 & ${AP}^{IOU=0.5}_{cat=x}$     & Average precision of $x$ class when IOU is 0.5\\
2 & ${AP}^{IOU=0.75}_{cat=x}$     & Average precision of $x$ class when IOU is 0.75\\ \hline
\end{tabular}
\end{table}

\begin{table}[!h]
\footnotesize
\centering
\caption{Test items (Mean Average Precision) in the prediction stage.}
\begin{tabular}{cll}
\hline
  & Indicator & Explanation \\ \hline
1 & $mAP$          & Mean average precision            \\
2 & ${mAP}^{IOU=0.5}$  & Mean average precision when IOU is 0.5             \\
3 & ${mAP}^{IOU=0.75}$  & Mean average precision when IOU is 0.75           \\
4 & ${mAP}_{size\_type=x}$ & Mean average precision when size type is $x$   \\
5 & ${mAP}_{occlusion\_level=x}$ & Mean average precision when occlusion level is $x$\\
6 & ${mAP}_{truncation\_level=x}$ & Mean average precision when truncation level is $x$ \\
7 & ${mAP}_{orientation\_level=x}$ & Mean average precision when orientation level is $x$  \\
8 & ${mAP}_{photo\_time=x}$ & Mean average precision when photo time is $x$   \\
9 & ${mAP}_{weather=x}$ & Mean average precision when weather is $x$      \\ \hline
\end{tabular}
\end{table}

\begin{table}[!h]
\scriptsize
\centering
\caption{Test items (Mean Average Recall) in the prediction stage.}
\begin{tabular}{cll}
\hline
  & Indicator & Explanation \\ \hline
1 & ${mAR}^{k}$  & when the maximum number of detections is $k$ \\
2 & ${mAR}^{k}_{IOU=0.5}$  & when IOU is 0.5 and the maximum number of detections is $k$ \\
3 & ${mAR}^{k}_{IOU=0.75}$  & when IOU is 0.75 and the maximum number of detections is $k$  \\
4 & ${mAR}^{k}_{size\_type=x}$  & when size type is $x$ and the maximum number of detections is $k$   \\
5 & ${mAR}^{k}_{occlusion\_level=x}$  & when occlusion level is $x$ and the maximum number of detections is $k$  \\
6 & ${mAR}^{k}_{truncation\_level=x}$  & when truncation level is $x$ and the maximum number of detections is $k$  \\
7 & ${mAR}^{k}_{orientation\_level=x}$  & when orientation level is $x$ and the maximum number of detections is $k$   \\
8 & ${mAR}^{k}_{photo\_time=x}$  & when photo time is $x$ and the maximum number of detections is $k$    \\
9 & ${mAR}^{k}_{weather=x}$  & when weather is $x$ and the maximum number of detections is $k$    \\  \hline
\end{tabular}
\end{table}

\begin{table}[!h]
\footnotesize
\centering
\caption{Test items (others) in the prediction stage.}
\begin{tabular}{clc}
\hline
  & Indicator & Explanation \\ \hline
1 & ${FPS}_{env=x}$  & FPS when test environment is $x$ in the prediction stage    \\
2 & Precision-Recall Curve Graph  &    \\
3 & Convergence Curve Graph  &    \\ \hline
\end{tabular}
\end{table}

\subsection{Classes and Annotations}
Table \ref{tab:cap3} shows the categories included in our dataset. The objects contained in our dataset can be roughly divided into four categories, including traffic related, human related, clothing, and appendage. Each category contains many subclasses, as shown in Table \ref{tab:cap3}.

In addition, the dataset indicates the location, size, occlusion ratio, truncation ratio, illumination and other attributes of the objects of the categories in Table \ref{tab:cap3}.

The position and size of the object are represented by the center, width and height of the object. Objects are divided into small objects, medium objects and large objects according to their width. As shown in the sub table on the left of Table \ref{tab:cap4}. Occlusion refers to natural continuous occlusion. Occlusion is divided into no occlusion, mild occlusion and severe occlusion according to its proportion. As shown in the sub table on the middle of Table \ref{tab:cap4}. Truncation refers to the proportion of the object beyond the frame. The truncation ratio is divided into no truncation, mild truncation and severe truncation. As shown in the sub table on the right of Table \ref{tab:cap4}.

In addition, as shown in Table \ref{tab:cap5}, we also represent other attributes, such as orientation, shooting period, and weather. Orientation includes front, side, oblique side, and back. Shooting period includes day, night, and early morning/evening. Weather includes sunny, overcast, rain, snow, and fog.

\subsection{Dataset Splits}
The dataset is divided into training set (including validation set) and test set (the test set does not provide annotation).

\section{Experiment}

The characteristic of our dataset is that it contains different scenes and has diversity in object detection properties, which can be used for research on different detection properties of existing object detection algorithms. We conduct experiments on different scenarios and different detection properties respectively. For different scenarios, we study the performance of different models on our dataset. For different detection properties, we study the effect of domain transfer, different occlusion ratios of objects, and different orientations.

\subsection{Multiple Object Detection Scenarios}
We evaluate the performance of different object detection models on intelligent transportation, intelligent surveillance, and drone scenarios, respectively. The models used for experiments include the lightweight model YOLO-v4\cite{bochkovskiy2020yolov4}, the commonly used CenterNet\cite{duan2019centernet}, Faster R-CNN\cite{ren2017faster}, and Transformer-based model DETR\cite{carion2020detr} and Deformable-DETR\cite{zhu2021deformable}.

\textbf{Experimental setting.} We trained YOLOv4 and YOLOv4-Tiny by SGD optimizer for 250 and 600 epochs, and the initial learning rates of models in traffic, surveillance, and drones scenarios are all 0.01. We trained CenterNet by Adam optimizer for 300 epochs, and the initial learning rates of models in traffic, surveillance, and drones scenarios are all 1.25e-4. The Faster R-CNN is trained with MMDetection\cite{chen2019mmdetection} which is an open-source object detection toolbox. We trained the Faster R-CNN by SGD optimizer for 24 epochs, and the initial learning rates of models in traffic, surveillance, and drones scenarios are set to be 0.05, 0.05, and 0.002 respectively. We trained the DETR for both three scenarios by AdamW optimizer for 300 epochs with a decay rate of 0.0001 and a batch size of 2 for each GPU. the DETR trained for both three scenarios by AdamW optimizer for 300 epochs with a decay rate of 0.0001 and a batch size of 2 for each GPU. The Deformable-DETR models of three scenarios are trained by AdamW optimizer with a decay rate of 0.0001 and a batch size of 2 too, and the training epochs are both set to 60. The backbone in both Faster R-CNN, DETR, and Deformable-DETR is ResNet-50.

Table {\color[rgb]{0.00,0.50,1.00} \ref{i_transportation}}, Table {\color[rgb]{0.00,0.50,1.00} \ref{i_surveillance}}, and Table {\color[rgb]{0.00,0.50,1.00} \ref{i_drone}} are the results of the trained models on the test sets of three scenarios: intelligent transportation, intelligent surveillance, and drones. We report the mean average precision (mAP) and mean average recall (mAR) for intersection over unions (IoUs) in [0.5:0.95], the APs at IoU=0.5 and 0.75, and the APs and ARs for small, medium and large objects.

\subsection{Qualitative Results}
\begin{figure*}[!h]
\begin{center}
\includegraphics[width=\textwidth]{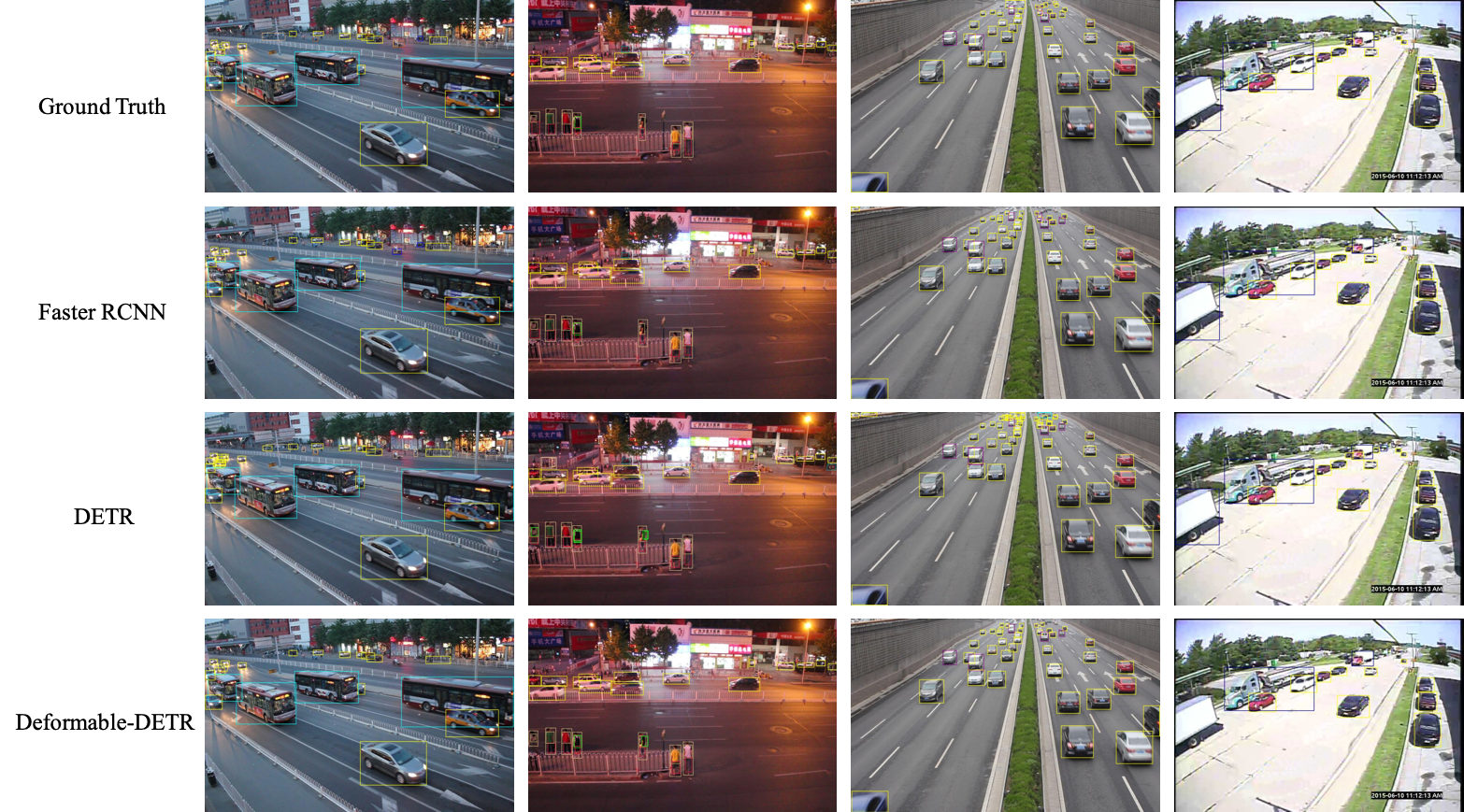}
\end{center}
   \caption{Qualitative results of Faster R-CNN on intelligent transportation scene. The first row shows the ground truth, the second row is the detection results of Faster R-CNN, the third row is detection results of DETR, and the fourth row is detection results of Deformable-DETR.}
\label{trans_det}
\end{figure*}

\begin{figure*}[!h]
\begin{center}
\includegraphics[width=\textwidth]{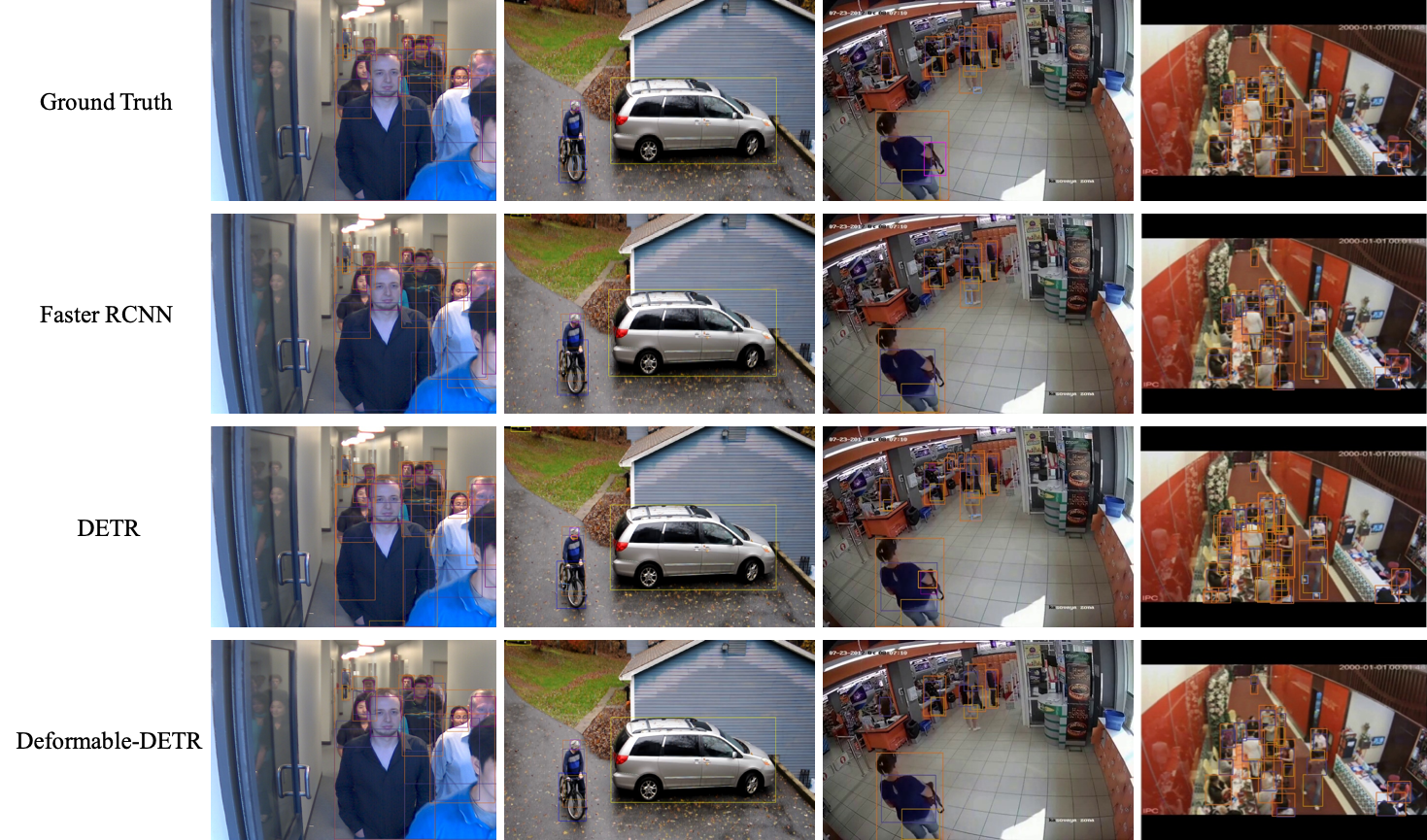}
\end{center}
   \caption{Qualitative results of Faster R-CNN on intelligent surveillance scene. The first row shows the ground truth, the second row is the detection results of Faster R-CNN, the third row is detection results of DETR, and the fourth row is detection results of Deformable-DETR.}
\label{sur_det}
\end{figure*}

\begin{figure*}[!h]
\begin{center}
\includegraphics[width=\textwidth]{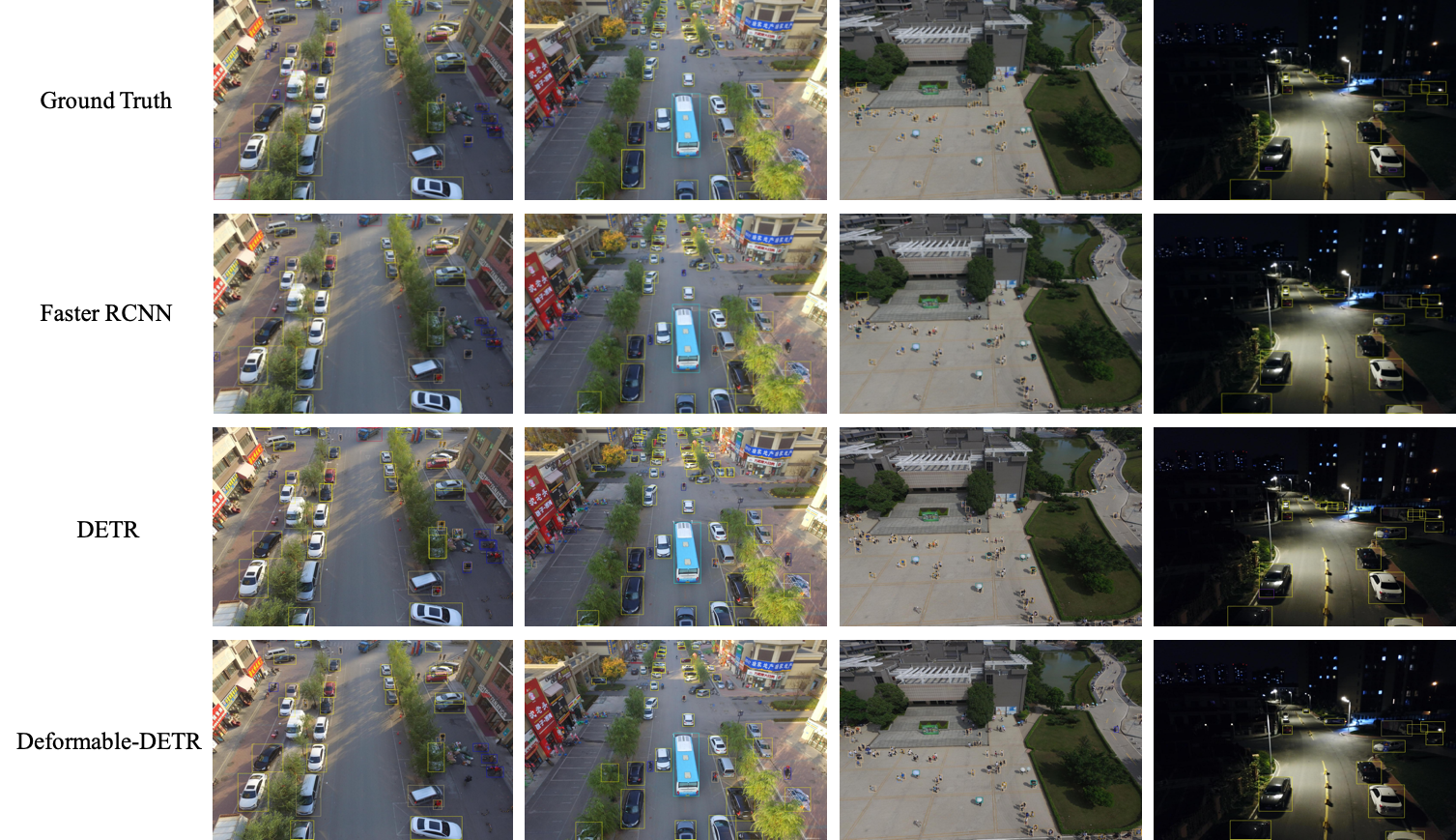}
\end{center}
   \caption{Qualitative results of Faster R-CNN on drone scene. The first row shows the ground truth, the second row is the detection results of Faster R-CNN, the third row is detection results of DETR, and the fourth row is detection results of Deformable-DETR.}
\label{drone_det}
\end{figure*}

Figure {\color[rgb]{0.00,0.50,1.00} \ref{trans_det}}, Figure {\color[rgb]{0.00,0.50,1.00} \ref{sur_det}} and Figure {\color[rgb]{0.00,0.50,1.00} \ref{drone_det}} show the visualization of the detection results of different object detection models in intelligent transportation, intelligent surveillance and drone scenarios. The first row in these figures is the ground truth of the objects, the second row in these figures is the detection results of Faster R-CNN, the third row in these figures is the detection results of DETR, and the last row in these figures is the detection results of Deformable-DETR.

As can be seen from the Figure {\color[rgb]{0.00,0.50,1.00} \ref{trans_det}} that the intelligent transportation scene in our data set contains images during the day, at night and under different lighting conditions, and the different models trained can accurately detect objects such as pedestrians, bags and vehicles of different sizes and types. Figure {\color[rgb]{0.00,0.50,1.00} \ref{sur_det}} shows that different models can accurately detect objects of different sizes and densities in security scenes with different angle cameras, such as top view and front view. Figure {\color[rgb]{0.00,0.50,1.00} \ref{drone_det}} shows that training different models in the UAV perspective scene can accurately detect small objects and objects at night.

\section{Empirical Setup}
\subsection{Learning Stage}
The test environment in the learning stage includes three environments based on GPU, AI chip and CPU. The specific parameters are as follows.

\subsection{Prediction Stage}
The test environment in the prediction stage includes three environments based on GPU, AI chip and FPGA. The specific parameters are as follows.

\section{Evaluation}
In the learning stage, we use Convergence Time (CT) and Epochs to evaluate the train. Convergence Time means the time from start of learning to convergence. Epochs means the total number of cycles from start of learning to convergence.

In the prediction stage, we adopt the following evaluation index.

\textbf{Average Precision} is also used to evaluate detection. For a given class, we compute the precision and recall for this class at different class confidence thresholds (from 0.0 to 1.0). And we can get the recall-precision curve with the x-axis as the recall rate and the y-axis as the precision. Then, we remove outliers from the curve in the following way, making the curve a monotonically decreasing curve. For each recall rate, the corresponding precision is taken as the maximum precision of all recall rates greater than or equal to the recall rate. The area under the calculated recall-precision curve is AP.

\textbf{Mean Average Precision} is mean precision over all classes as follows:
\begin{equation}
\begin{aligned}
      mAP = \frac{\sum_{i=0}^{N}{AP}_i}{N}.\\
\end{aligned}
\end{equation}

\textbf{Average Recall} is defined as:
\begin{equation}
\begin{aligned}
      AR_{max=k}^{IOU=0.5:0.05:0.95} = \frac{\sum_{x}R_{max=k}^{IOU=x}}{11},\\
\end{aligned}
\end{equation}
$R_{max=k}^{IOU=x}$ means the recall rate when the IOU is $x$ and the maximum number of detection examples of an image is $k$. The values of $k$ are 1, 10, 100, 500. $AR_{max=k}^{IOU=0.5:0.05:0.95}$ is abbreviated as ${AR}^k$ in the following text.

\textbf{Mean Average Precision} is defined as:
\begin{equation}
\begin{aligned}
      {mAR}^k=\frac{\sum_{i=0}^{N}{AR}^{k}_{i}}{N}.\\
\end{aligned}
\end{equation}

\textbf{Frame per Second} means the number of frames that the detection algorithm can process in 1 second.


\section{Test Report and Ranking}
After completing the algorithm test, this benchmark will provide a test report and the ranking of the algorithm among all participating algorithms. The test report contains the test results for the following test items.

\section{Test Portal}\label{sectestportal}
This benchmark provides a testing portal in the form of a website. Through this portal, users only need to upload the container of the test program that conforms to the specifications of Chapter 9, then they can complete the test of each test item in Chapter 7 in each test environment in Chapter 6, and get the test report and test report of Chapter 8. ranking.
\section{Test Method}
The following test methods are based on the Test Portal implementation in Section \ref{sectestportal}. The test pipeline is shown in below.


\subsection{Login to the Test Portal}
The user first registers an account in the test portal, logs in with the account and goes to personal page.

\subsection{Upload Test Program Container}
The user uploads the container containing the learning program and the test program in accordance on the personal page.

\subsection{Create A New Test Task}
The user fills in the relevant information of the test task on the New Test Task page and submits the test task. After the test task is submitted, the test portal will test the algorithm according to the test items in each test environment.

\subsection{Get Test Results}
After the submitted algorithm test task is completed, the user can view the test report and ranking of the algorithm.

\section{Conclusion}
In this work, we presented a new benchmark for urban scenes, a large-scale dataset with extensive annotations in semantic urban scene understanding. We built a large-scale object detection benchmark and conduct extensive evaluation of existing methods. We hope our work can support future studies on this important direction.

\clearpage
%
%
\bibliographystyle{splncs04}
\bibliography{egbib}
\end{document}